\documentclass[conference]{IEEEtran}
\IEEEoverridecommandlockouts
\usepackage{cite}
\usepackage{amsmath,amssymb,amsfonts}
\usepackage{algorithmic}
\usepackage{algorithm}
\usepackage{array}
\usepackage{textcomp}
\usepackage{url}
\usepackage{graphicx}
\usepackage{multirow}
\usepackage{booktabs}
\usepackage{threeparttable}
\usepackage{bm}
\usepackage[table,xcdraw]{xcolor}
\usepackage{tikz}
\usepackage{adjustbox}
\usepackage{etoolbox}
\def\BibTeX{{\rm B\kern-.05em{\sc i\kern-.025em b}\kern-.08em
		T\kern-.1667em\lower.7ex\hbox{E}\kern-.125emX}}

\makeatletter
\g@addto@macro\normalsize{%
	\setlength{\abovedisplayskip}{2.5pt plus 1pt minus 1pt}%
	\setlength{\belowdisplayskip}{2.5pt plus 1pt minus 1pt}%
	\setlength{\abovedisplayshortskip}{1.5pt plus 1pt minus 1pt}%
	\setlength{\belowdisplayshortskip}{1.5pt plus 1pt minus 1pt}%
}

\def\subsection{\@startsection{subsection}{2}{\z@}{2.2ex plus 0.6ex minus 0.2ex}%
	{0.95ex plus 0.25ex minus 0.1ex}{\normalfont\normalsize\itshape}}%
\makeatother

\newcommand{\redta}[1]{#1}

\newif\ifrevised
\revisedtrue        

\ifrevised
\newcommand{\revmwl}[1]{{\color{black}#1}}

\newcommand{\revlzh}[1]{#1}

\newif\ifshortened
\shortenedtrue        
\DeclareRobustCommand{\revshort}[2]{\ifshortened{\color{black}#2}\else#1\fi}
\DeclareRobustCommand{\revkeep}[2]{#1}

\else

\newcommand{\revshort}[2]{#2}
\newcommand{\revkeep}[2]{#1}
\fi
\DeclareRobustCommand{\revfinal}[1]{{\color{black}#1}}

\begin{document}
	
	\title{Rapid and Safe Trajectory Planning over Diverse Scenes through Diffusion Composition}
	\author{
		\IEEEauthorblockN{Wule Mao$^{1}$, Zhouheng Li$^{1}$, Yunhao Luo$^{2}$, Fangguo Zhao$^{1}$, Lei Xie$^{1,\dagger}$}
		\IEEEauthorblockA{$^{1}$Zhejiang University, Hangzhou, China, 
			$^{2}$University of Michigan, MI, USA\\
			$^{\dagger}$Corresponding author: leix@iipc.zju.edu.cn}
		\vspace{-30pt}
	}

	\maketitle

	\begin{abstract}
		\revkeep{Achieving safe, efficient, and kinematically feasible planning in dynamic environments remains a significant challenge, as planners must simultaneously handle moving obstacles, sensor uncertainty, and strict motion constraints. To address this problem,  we propose an energy-parameterized diffusion planning framework that learns a conservative energy field to realize safe and stable generalization across diverse scenarios. The energy-parameterized diffusion formulation enables flexible integration of multiple constraints, allowing the planner to generalize to previously unseen environments without retraining. To ensure real-time safety during deployment, we further incorporate a lightweight safety filter that enforces safety and kinematic feasibility constraints in real-time. Additionally, we develop a scene-agnostic, MPC-based data generation pipeline to produce large-scale, dynamically feasible training trajectories. In simulation, the proposed method achieves real-time performance with a mean planning time of 0.21s and a low planning failure rate of 0.57\%. Real-world experiments on the F1TENTH platform further validate the effectiveness of the proposed framework. Under sensor uncertainty in previously unseen dynamic environments, the planner consistently generates collision-free trajectories, which remain safe after being tracked by a simple controller, maintaining a mean obstacle clearance of 0.26 m, demonstrating strong robustness and practical applicability.}{Safe real-time planning in dynamic scenes must handle moving obstacles, sensor uncertainty, and vehicle kinematics. We propose RSTP, an energy-parameterized diffusion planner that learns a conservative energy field for stable test-time composition across diverse scenes. The framework integrates multiple constraints without retraining and uses a lightweight safety filter to select trajectories satisfying safety and kinematic limits. A scene-agnostic MPC pipeline generates dynamically feasible demonstrations. In simulation, RSTP achieves 0.21s mean planning time and 0.57\% failure rate. Real-world F1TENTH experiments further show collision-free tracking in unseen dynamic scenes under sensor uncertainty, with 0.26 m mean obstacle clearance.}
		Project page: \url{https://rstp-comp-diffuser.github.io}.
		
	\end{abstract}
	
	
	\begin{figure*}[!t]
		\centering
		\includegraphics[width=0.95\textwidth]{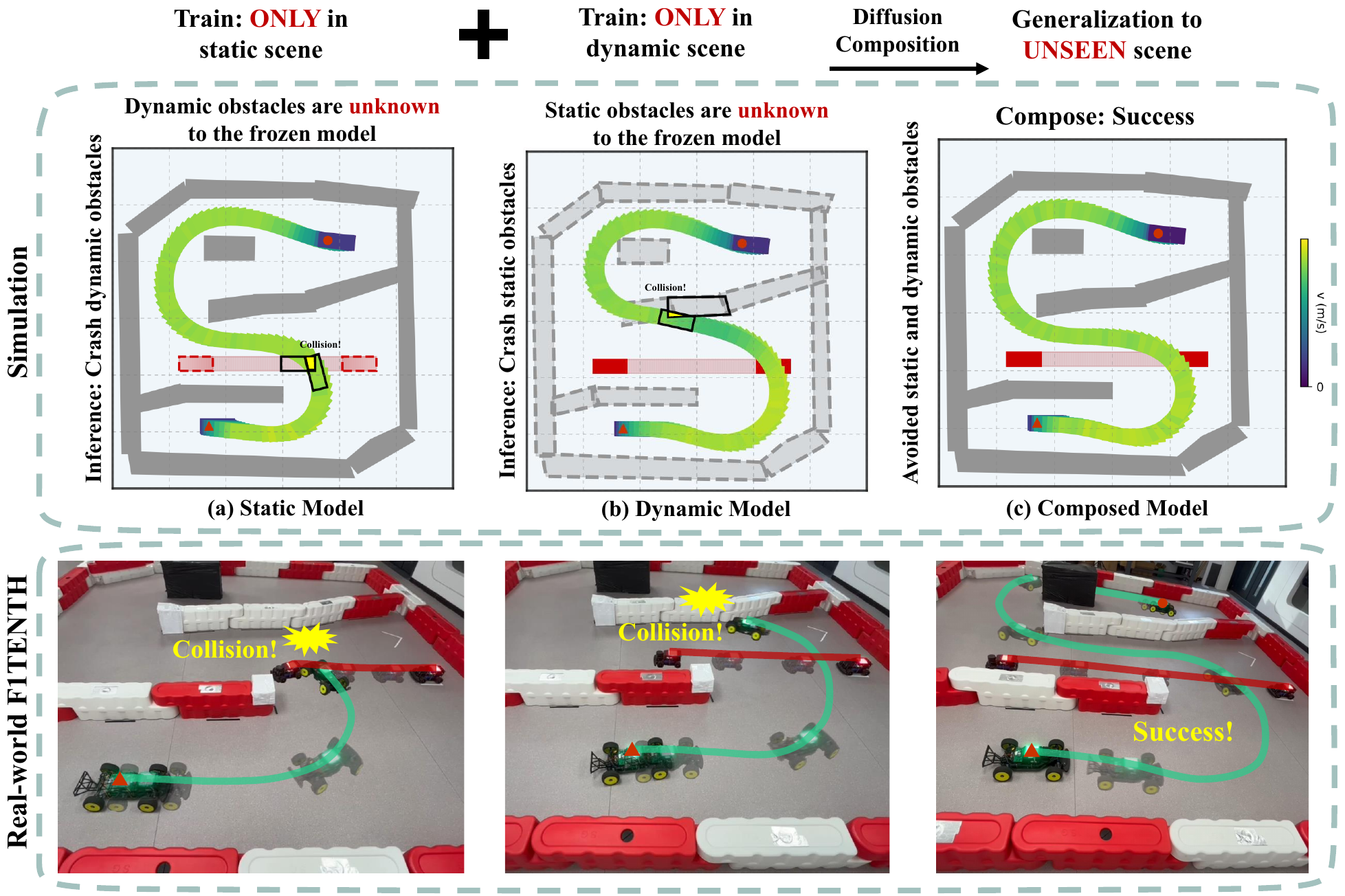}
		\caption{\textbf{Diffusion Composition Enables Efficient, Safe Planning with Practical Real-world Performance.} An individual diffusion model cannot ensure safe trajectory planning in out-of-distribution scenarios, whereas composing multiple energy-based diffusion models can achieve test-time generalization. Dashed boxes indicate obstacles that do not exist during training. Validation on the F1TENTH platform shows that trajectories planned by the composed diffusion model offer excellent safety while maintaining computational efficiency, demonstrating effectiveness for practical real-world applications.}
		\label{fig:vis_abstract}
		\vspace{-5pt}
	\end{figure*}
	
	\section{Introduction}

	\revkeep{Trajectory planning in unknown environments with both static and dynamic obstacles remains a central challenge in local planning~\cite{han2023efficient,liRapidIterativeTrajectory}. A practical planner must generate collision-free trajectories in real time while satisfying kinematic constraints, making local planning computationally demanding in complex scenes~\cite{liRapidIterativeTrajectory,7402333,0278364914528132}. Search or sampling-based methods are widely used for planning~\cite{dolgov2008practical,karaman2011sampling}. However, their computational cost increases significantly with scene complexity, limiting real-time reliability~\cite{dolgov2008practical}. Moreover, vehicle kinematic constraints are difficult to incorporate directly during search or sampling~\cite{sakcak2019sampling,dolgov2008practical}, often requiring post-processing optimization. Optimal control approaches, such as Model Predictive Control (MPC), explicitly enforce system dynamics and constraints to produce feasible trajectories~\cite{zhang2020optimization,yamaguchi2021model}. Yet they usually depend on heuristic initializations and may suffer from local minima, functioning mainly as refinement modules rather than standalone local planners~\cite{zhang2020optimization,liRapidIterativeTrajectory}.}{Trajectory planning among static and dynamic obstacles requires real-time collision avoidance under vehicle kinematics~\cite{han2023efficient,liRapidIterativeTrajectory}. Search and sampling methods scale poorly with scene complexity and often enforce kinematics only through post-processing~\cite{dolgov2008practical,sakcak2019sampling}. Optimal-control planners such as MPC impose dynamics explicitly, but they rely on warm starts and may converge to local minima, making them better suited as refiners than standalone local planners~\cite{zhang2020optimization,liRapidIterativeTrajectory}.}
	
	\revkeep{Imitation learning provides an alternative paradigm by directly mapping sensory observations to executable trajectories~\cite{chi2023diffusion,zheng2025diffusionbased,tan2025flow}. Diffusion models, in particular, have demonstrated strong generative capability and the ability to capture multi-modal trajectory distributions~\cite{chi2023diffusion,janner2022planning}. However, their performance often degrades in previously unseen dynamic environments due to distributional shifts, limiting robustness and generalization in real-world deployment~\cite{du2023reduce}.}{Imitation learning maps observations directly to executable trajectories. Diffusion models capture multimodal behaviors, but distribution shifts in unseen dynamic scenes can degrade safety and generalization~\cite{chi2023diffusion,du2023reduce}.}
	
	\revkeep{To address these challenges, we propose the \textbf{R}apid and \textbf{S}afe \textbf{T}rajectory \textbf{P}lanning (\textbf{RSTP}) framework, a conservative diffusion-based approach for local planning in dynamic environments. The proposed method generates collision-free and kinematically feasible trajectories in unseen complex scenes, achieving improved planning efficiency while preserving safety and feasibility. In summary, the proposed RSTP framework has the following contributions:}{To address these challenges, we propose \textbf{R}apid and \textbf{S}afe \textbf{T}rajectory \textbf{P}lanning (\textbf{RSTP}), a conservative diffusion planner for efficient, safe, and kinematically feasible local planning in unseen dynamic scenes. Its main contributions are:}
	\begin{enumerate}
		\item \redta{\textbf{\revlzh{Conservative} Diffusion Composition for Flexible Planning across Diverse Scenes:}}
		\revkeep{\revlzh{We introduce an energy-parameterized diffusion trajectory planner that enables test-time composition, allowing safe generalization to previously unseen scenarios without additional training, while maintaining kinematic feasibility under distribution shifts.}}{We introduce an energy-parameterized diffusion planner that supports test-time composition and safe zero-shot generalization while preserving kinematic feasibility.}

		\item \redta{\textbf{Rapid and Safe Diffusion Planning from Localized State Estimates:}}
		\revkeep{\redta{Leveraging low-dimensional vehicle state representations, RSTP achieves a mean planning time of \textbf{$0.21\,\text{s}$} on consumer-grade hardware. A lightweight rule-based safety filter selects the most feasible trajectory from a batch of sampled candidates, enforcing safety and kinematic constraints and reducing the planning failure rate to \textbf{0.57\%}.}}{Using low-dimensional vehicle states, RSTP plans in \textbf{$0.21\,\text{s}$} on consumer hardware; a rule-based safety filter enforces safety and kinematic limits and reduces failures to \textbf{0.57\%}.}

		\item \redta{\textbf{Validation on a Real-World Scaled F1TENTH Vehicle Platform:}}
		\revkeep{\redta{We validate the proposed approach on an F1TENTH vehicle using server-side inference and onboard pure-pursuit tracking. The vehicle accurately tracks the planned trajectories without collisions while maintaining a safe mean obstacle clearance of \textbf{0.26\,m}, demonstrating kinematic feasibility and practical applicability for real-world applications.}}{F1TENTH experiments with server-side inference and pure-pursuit tracking achieve collision-free execution with \textbf{0.26\,m} mean clearance.}
		
	\end{enumerate}

	\revkeep{Section \ref{sec:Preliminaries} presents the energy-parameterized diffusion formulation. \revfinal{Section \ref{sec:Method} describes the proposed framework, including diffusion composition, the safety filter, and the MPC-based dataset generation process.} Section \ref{sec:Exp} reports simulation results and experimental validation on the F1TENTH platform. Finally, Section \ref{sec:Conclusion} concludes the paper by summarizing the main findings.}{The remainder presents the formulation, method, experiments, and conclusion.}
	
	\section{Related Work}
	\revkeep{This section reviews related work on trajectory planning and generative modeling, focusing on \revfinal{hierarchical planning approaches} and energy-based diffusion models relevant to our framework.}{We review hierarchical planning and energy-based diffusion models most relevant to RSTP.}
	\vspace{-10pt}
	
	\subsection{Hierarchical Planning Approach}
	\revkeep{Traditional trajectory planning in complex environments typically follows a two-stage framework: a collision-free path is generated and then refined through optimization. Search~\cite{liu2017global} and sampling-based~\cite{karaman2011sampling,wu2021st} methods become computationally expensive as scene complexity increases and struggle to incorporate kinematic constraints directly while ensuring success rate~\cite{sakcak2019sampling,dolgov2008practical}. \revfinal{MPC-based methods can enforce system dynamics and safety constraints but often rely on heuristic initializations}~\cite{zhang2020optimization}, are prone to local minima, and incur high computational costs in cluttered environments. Model Predictive Contouring Control  (MPCC) is also a promising approach, but it similarly requires path initialization~\cite{lyons2023curvature,li2024data,li2025reduce,li6127037evo}. Model Predictive Path Integral (MPPI) is an efficient method for stochastic optimal control~\cite{honda2025model,zhao2026vision}. However, it typically samples from a unimodal proposal distribution~\cite{honda2024stein}, which makes it prone to local minima and limits its ability to escape suboptimal solutions.}{Classical planners often generate a collision-free path and refine it by optimization. Search and sampling methods are costly in complex scenes and weakly handle kinematics~\cite{liu2017global,karaman2011sampling,wu2021st,sakcak2019sampling}. MPC and MPCC impose dynamics but require path initialization and may converge locally~\cite{zhang2020optimization,lyons2023curvature,li2024data}. MPPI is efficient, yet its unimodal proposal can limit escape from local minima~\cite{honda2025model,honda2024stein}.}
	
	\subsection{Energy-based Diffusion Model}
	\revkeep{Energy-based models (EBMs) offer attractive properties, such as conservative gradient fields~\cite{chao2022investigating} and compositionality~\cite{du2023reduce}, enabling principled constraint integration at test time. However, directly learning explicit energy functions for trajectory behavior cloning is often unstable~\cite{chi2023diffusion}.
		Diffusion models can be interpreted as learning the gradient field of an implicit energy function~\cite{song2020denoising}. Owing to their strong generative capacity and robustness in trajectory modeling~\cite{chi2023diffusion}, they have been widely adopted for motion generation and planning~\cite{ma2025constraint,mizuta2024cobl,christopher2024constrained,wang2024poco}. Building on this connection, we leverage diffusion models to approximate the energy gradient, providing a practical and stable approach to training implicit energy-parameterized diffusion models for trajectory generation.}{EBMs provide conservative fields and compositionality for test-time constraint integration~\cite{chao2022investigating,du2023reduce}, but explicit energy learning can be unstable~\cite{chi2023diffusion}. Since diffusion models estimate gradients of implicit energies~\cite{song2020denoising}, they offer a stable route to energy-parameterized trajectory generation and planning~\cite{ma2025constraint,mizuta2024cobl,christopher2024constrained}.}

	\section{Preliminaries}\label{sec:Preliminaries}
	
	\revshort{In this work, trajectory planning is formulated as sampling over a learned energy landscape using an energy-parameterized diffusion model to represent its gradient.}{We formulate planning as sampling over a learned energy landscape via an energy-parameterized diffusion model.}
	\vspace{-5pt}
	\subsection{Energy-parameterized Diffusion Representation}
	Let $\boldsymbol{\tau} = [\mathbf{s}_1, \dots, \mathbf{s}_L] \in \mathbb{R}^{n \times L}$ denote a discrete trajectory with horizon $L$, and let $\mathcal{C}$ represent the planning conditions. We model the conditional trajectory distribution using an energy-based formulation~\cite{du2023reduce}:
	$
	p_\theta(\boldsymbol{\tau} \mid \mathcal{C})
	\propto
	\exp\big(-E_\theta(\boldsymbol{\tau}, \mathcal{C})\big),
	$
	where $E_\theta(\boldsymbol{\tau}, \mathcal{C})$ is a learnable neural network defining the energy landscape. Low-energy trajectories correspond to feasible and collision-free motions.
	From a score-based perspective, diffusion models learn the gradient field of the data distribution~\cite{song2020denoising}:
	$
	\epsilon_\theta(\boldsymbol{\tau}, \mathcal{C})
	\propto
	-\nabla_{\boldsymbol{\tau}} \log p_\theta(\boldsymbol{\tau} \mid \mathcal{C}).
	$
	Substituting the energy-based density yields
	$
	\epsilon_\theta(\boldsymbol{\tau}, \mathcal{C})
	\propto
	\nabla_{\boldsymbol{\tau}} E_\theta(\boldsymbol{\tau}, \mathcal{C}).
	$
	\revshort{Therefore, the diffusion noise can be interpreted as approximating the gradient of the underlying energy function~\cite{song2020denoising}.}{Thus, diffusion noise approximates the energy gradient~\cite{song2020denoising}.}
	
	During the reverse diffusion process, the trajectory becomes time-dependent and is indexed by the diffusion step $t$, denoted as $\boldsymbol{\tau}^t$. Therefore, the reverse diffusion update can be written as
	$$
	\boldsymbol{\tau}^{t-1}
	=
	\boldsymbol{\tau}^{t}
	-
	\lambda \nabla_{\boldsymbol{\tau}^t} E_\theta(\boldsymbol{\tau}^{t}, t, \mathcal{C})
	+
	\sigma_t \boldsymbol{\xi},
	$$
	where $\boldsymbol{\xi} \sim \mathcal{N}(\mathbf{0}, \mathbf{I})$, $\lambda$ denotes the gradient step size, and $\sigma_t$ is the noise scale at step $t$.
	\revshort{This perspective interprets diffusion-based trajectory generation as energy minimization over the learned energy landscape.}{Diffusion planning can thus be viewed as energy minimization.}
	
	
	\subsection{Conservative Property of Energy-Based Models}

	\begin{figure*}[!t]
		\centering
		\includegraphics[width=7.00in]{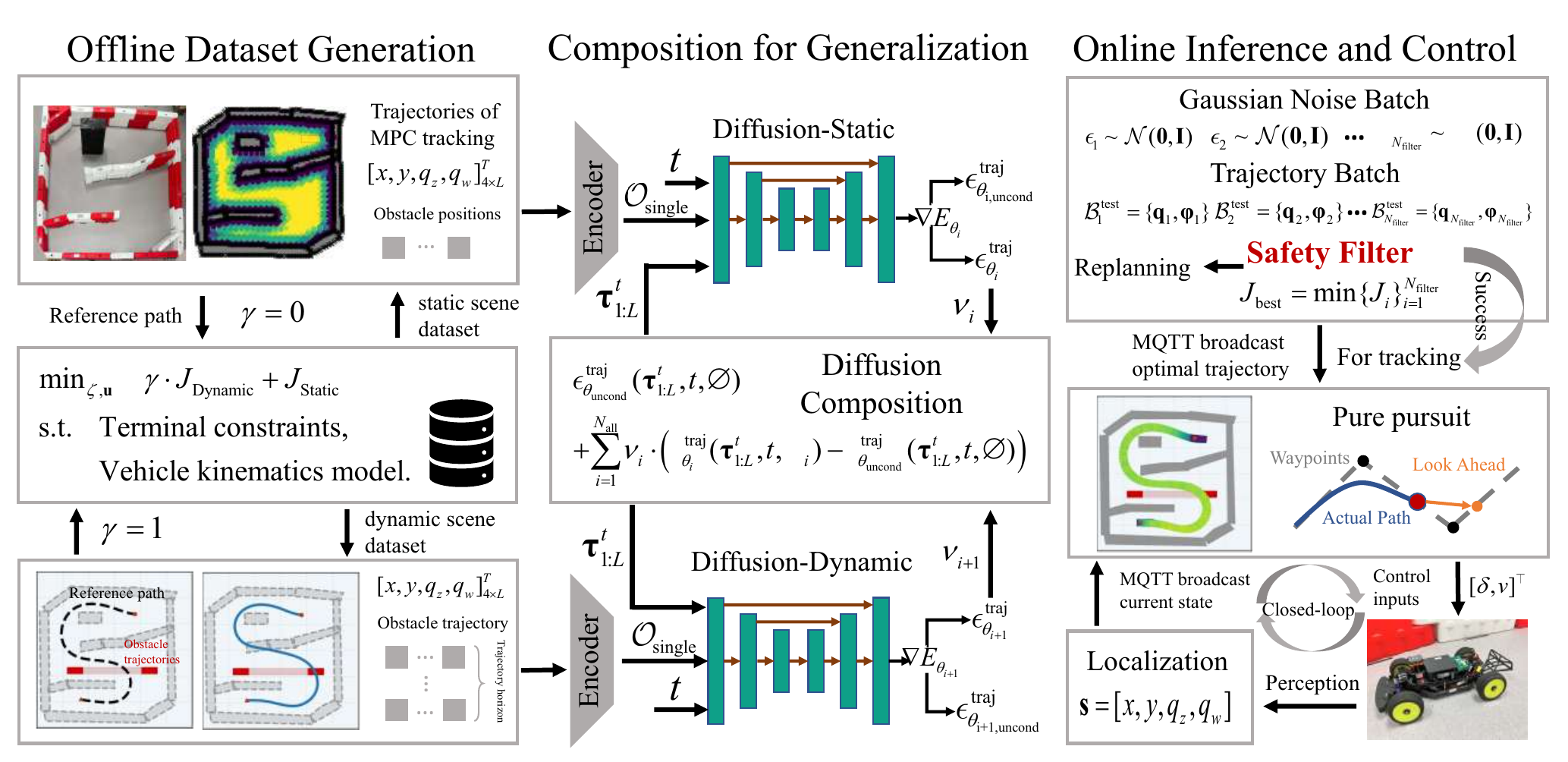}
		\caption{\revkeep{\textbf{The Overall Framework of the Proposed Rapid and Safe Trajectory Planning (RSTP) Method.} \textit{Offline Dataset Generation} (Left): MPC-based methods provide kinematically feasible trajectory datasets for training. \textit{Diffusion Composition} (Middle): individual \revfinal{energy-parameterized} diffusion models can be flexibly composed to tackle novel scenarios not covered in the training data. \textit{Online Inference and Control} (Right): The ego vehicle performs real-time inference from its current pose, and the safety filter selects the optimal trajectory per cycle for tracking.}{\textbf{RSTP framework.} MPC generates feasible demonstrations, energy-parameterized diffusion models are composed for unseen scenes, and online inference uses a safety filter to select a trackable trajectory.}}
		\label{fig:pipeline}
	\end{figure*}	
	
	An important property of energy-based models (EBMs) is that their associated score field is conservative~\cite{chao2022investigating, salimans2021should}. Since the diffusion noise $\epsilon_\theta$ corresponds to the gradient of a scalar potential function $E_\theta$, i.e.,
	$
	\epsilon_\theta(\boldsymbol{\tau}, t, \mathcal{C}) = \nabla_{\boldsymbol{\tau}} E_\theta(\boldsymbol{\tau}, t, \mathcal{C}),
	$
	the resulting vector field is necessarily curl-free~\cite{chao2022investigating}. Therefore, $\epsilon_\theta$ is conservative if and only if
	$
	\nabla_{\boldsymbol{\tau}} \times \epsilon_\theta(\boldsymbol{\tau}, t, \mathcal{C}) = \mathbf{0},
	$
	which guarantees the existence of a globally defined scalar potential function~\cite{chen2014stochastic}.
	Equivalently, the energy-based formulation satisfies the symmetry condition on mixed partial derivatives:
	$
	\frac{\partial \epsilon_i}{\partial \tau_j}
	=
	\frac{\partial \epsilon_j}{\partial \tau_i},
	\forall i,j \in \{1, \dots, n L\},
	$
	where 
	$
	\frac{\partial \epsilon_i}{\partial \tau_j}
	=
	\frac{\partial}{\partial \tau_j}
	\left(
	\frac{\partial E_\theta}{\partial \tau_i}
	\right)
	=
	\frac{\partial^2 E_\theta}{\partial \tau_j \partial \tau_i},
	$
	and the equality of mixed partial derivatives ensures:
	$$
	\frac{\partial^2 E_\theta}{\partial \tau_j \partial \tau_i}
	=
	\frac{\partial^2 E_\theta}{\partial \tau_i \partial \tau_j}.
	$$
	\revkeep{When this condition holds, the learned vector field corresponds to a globally consistent energy landscape.
		It prevents cyclic or inconsistent gradient behaviors. This conservative structure is particularly desirable for trajectory planning. Although score fields support additive composition~\cite{chao2022investigating,du2023reduce}, directly combining multiple diffusers can introduce non-zero curl and lead to instability~\cite{chi2023diffusion}. Energy parameterization mitigates this issue by enforcing a conservative field, making energy-parameterized diffusion necessary for stable composition.}{This yields a globally consistent energy landscape and avoids cyclic gradients. }
	\vspace{-5pt}
	
	\section{The Proposed Rapid and Safe Trajectory Planning Method} \label{sec:Method}
	
	\revshort{Fig.~\ref{fig:pipeline} illustrates the proposed RSTP pipeline. This section details the energy-parameterized diffusion model for generalization to unseen scenes through composition of individually trained models. It also describes the safety filter applied at test time to ensure safety.}{Fig.~\ref{fig:pipeline} shows the RSTP pipeline, including diffusion composition and test-time safety filtering.}
	
	\subsection{Energy-Parameterized Diffusion Composition}
	\revkeep{Let $\boldsymbol{\tau}_{1:L} = [x, y, \cos(\varphi), \sin(\varphi)] \in \mathbb{R}^{4 \times L}$ denote a discrete trajectory, where $(x,y)$ are Cartesian coordinates \revfinal{and} $\varphi$ is the heading angle. Velocities are implicitly encoded through a fixed time interval $\Delta t$. We model the conditional trajectory distribution under obstacle condition $\mathcal{O}_i$ using an energy-based formulation:}{Let $\boldsymbol{\tau}_{1:L} = [x, y, \cos(\varphi), \sin(\varphi)] \in \mathbb{R}^{4 \times L}$ be a discrete trajectory with fixed interval $\Delta t$. Under obstacle condition $\mathcal{O}_i$,}
	$p^{\text{traj}}_{\theta_i}(\boldsymbol{\tau}_{1:L} \mid \mathcal{O}_i)
	\propto
	\exp\!\big(-E_{\theta_i}(\boldsymbol{\tau}_{1:L}, \mathcal{O}_i)\big),$
	where $E_{\theta_i}$ is a learned energy function. 
	

	\revkeep{Inspired by ~\cite{du2023reduce}, we adopt the implicit energy parameterization. The neural network outputs a vector field}{Following~\cite{du2023reduce}, the network outputs}
	$f_\theta(\boldsymbol{\tau}_{1:L}, t, \mathcal{O})$,
	\revkeep{and the implicit energy landscape is defined as:}{and defines}
	\begin{equation}
		E_\theta(\boldsymbol{\tau}_{1:L}, t, \mathcal{O})
		=
		\frac{1}{2}
		\left\|
		f_\theta(\boldsymbol{\tau}_{1:L}, t, \mathcal{O})
		\right\|^2,
	\end{equation}
	\revkeep{where $t$ is the diffusion step. \revfinal{The} energy gradient is:}{where $t$ is the diffusion step. Its gradient is}
	\begin{equation}
		\nabla_{\boldsymbol{\tau}_{1:L}}
		E_\theta(\boldsymbol{\tau}_{1:L}, t, \mathcal{O})
		=
		J_{f_\theta}^\top
		f_\theta(\boldsymbol{\tau}_{1:L}, t, \mathcal{O}).
	\end{equation}
	where $J_{f_\theta}$ denotes the Jacobian of $f_\theta$. 
	\revkeep{Therefore, the diffusion noise is:}{and the noise prediction is}
	$
	\epsilon^{\text{traj}}_\theta(\boldsymbol{\tau}_{1:L}, t, \mathcal{O})
	=
	\nabla_{\boldsymbol{\tau}_{1:L}}
	E_\theta(\boldsymbol{\tau}_{1:L}, t, \mathcal{O}).
	$
	\revkeep{Then diffusion sampling corresponds to gradient descent on the learned energy landscape. This formulation ensures that the learned vector field is conservative~\cite{chao2022investigating}, which is essential for stable and compositional modeling.}{Thus, sampling performs gradient descent on a conservative learned energy field~\cite{chao2022investigating}.}

	
	For unseen environments containing multiple obstacles, $\mathcal{O}_{\text{all}} = \{\mathcal{O}_1, \dots, \mathcal{O}_{N_{\text{all}}}\}$, we construct the composed distribution under the assumption of energy additivity. The total energy is defined as a weighted sum of individual energies:
	$$
	\begin{aligned}
		p^{\text{traj}}_{\theta_{\text{compose}}}
		(\boldsymbol{\tau}_{1:L} \mid \mathcal{O}_{\text{all}})
		&\propto
		\exp\left(
		-\sum_{i=1}^{N_{\text{all}}}
		E_{\theta_i}(\boldsymbol{\tau}_{1:L}, \mathcal{O}_i)
		\right)\\
		&\propto
		p^{\text{traj}}_{\theta_{\text{uncond}}}(\boldsymbol{\tau}_{1:L})
		\prod_{i=1}^{N_{\text{all}}}
		\left(
		\frac{
			p^{\text{traj}}_{\theta_i}(\boldsymbol{\tau}_{1:L} \mid \mathcal{O}_i)
		}{
			p^{\text{traj}}_{\theta_{\text{uncond}}}(\boldsymbol{\tau}_{1:L})
		}
		\right).
	\end{aligned}
	$$
	where $p^{\text{traj}}_{\theta_{\text{uncond}}}(\boldsymbol{\tau}_{1:L})$ is obtained by masking conditions. Substituting the energy-based parameterization yields the composed noise prediction, which can be written as:
	\begin{equation}
		\begin{aligned}
			\epsilon^{\text{traj}}_{\theta_{\text{compose}}}
			&(\boldsymbol{\tau}_{1:L}^t, t,
			\{\mathcal{O}_i\}_{i=1}^{N_{\text{all}}}) 
			=
			\epsilon^{\text{traj}}_{\theta_{\text{uncond}}}
			(\boldsymbol{\tau}_{1:L}^t, t, \emptyset)\\
			&+
			\sum_{i=1}^{N_{\text{all}}}
			\nu_i
			\left(
			\epsilon^{\text{traj}}_{\theta_i}
			(\boldsymbol{\tau}_{1:L}^t, t, \mathcal{O}_i)
			-
			\epsilon^{\text{traj}}_{\theta_{\text{uncond}}}
			(\boldsymbol{\tau}_{1:L}^t, t, \emptyset)
			\right).
		\end{aligned}\label{eq:compose}
	\end{equation}
	Here, $\nu_i \ge 0$ controls the contribution of each noise term.
	The composed score equals the unconditional prediction plus weighted conditional residuals, making diffusion composition equivalent to weighted energy aggregation and its sampling interpretable as gradient descent on the composed energy landscape. Unlike~\cite{luo2024potential}, we use a shared unconditional noise prediction when combining conditional residuals, which preserves a consistent base score and improves compositionality.
	
	\revkeep{In our framework, obstacle information is encoded as constraints via the Transformer architecture proposed in~\cite{torchvision2016}, which maps $\mathcal{O}_i$ into a latent conditioning representation that modulates the trajectory denoising process. The encoder adopts a 1D Vision Transformer that partitions the obstacle representation into patches and projects them into a sequence of embeddings with a learnable class token.}{Obstacle constraints are encoded by a 1D Vision Transformer~\cite{torchvision2016}, which embeds $\mathcal{O}_i$ to condition trajectory denoising.}

	\begingroup
	\setlength{\abovedisplayskip}{1pt}
	\setlength{\belowdisplayskip}{1pt}
	\setlength{\abovedisplayshortskip}{0pt}
	\setlength{\belowdisplayshortskip}{1pt}
	\setlength{\jot}{0pt}
	\setlength{\parskip}{0pt}
	\subsection{Safety Filter: Test-Time Trajectory Selection}
	
	\revkeep{To bridge the gap between stochastic generation and safe deployment, we perform post-sampling refinement. This process filters raw trajectory candidates based on kinematic limits and obstacle clearances. The overall procedure of the algorithm is in Algorithm~\ref{alg:RSTP}.}{Algorithm~\ref{alg:RSTP} refines the sampled batch and selects candidates that satisfy obstacle-clearance and kinematic limits before execution.}
	
	\subsubsection{Batch Trajectory Generation}
	
	\revfinal{For each planning cycle, the planner performs at most $N_{\text{retry}}$ replanning attempts. In each attempt, a batch of $N_{\text{filter}}$ candidate trajectories is generated for selection:}
	$$
	\mathcal{B}^{\text{test}}_{1:N_{\text{filter}}}
	=
	\left\{
	\mathbf{q}_i, \boldsymbol{\varphi}_i
	\right\}_{i=1}^{N_{\text{filter}}}
	$$
	\revkeep{where $\mathbf{q}_i \in \mathbb{R}^{2 \times L}$ denotes the 2D positions and $\boldsymbol{\varphi}_i \in \mathbb{R}^{1 \times L}$ denotes the yaw sequence.
		Each trajectory is evaluated independently using geometric, kinematic, and safety metrics.}{where $\mathbf{q}_i$ and $\boldsymbol{\varphi}_i$ are position and yaw sequences evaluated by geometric, kinematic, and safety metrics.}

	\subsubsection{Geometric and Kinematic Evaluation}

	The path length $\mathcal{L}_i$ is used to assess the motion efficiency, computed as:
	$$
	d_{i,j}
	=
	\|\mathbf{q}_{i,[:,j+1]} - \mathbf{q}_{i,[:,j]}\|_2,
	\qquad
	\mathcal{L}_i = \sum_j d_{i,j}.
	$$

	Velocity and acceleration are calculated as:
	$$
	v_{i,j}
	=
	\frac{d_{i,j}}{\Delta t},
	\qquad
	a_{i,j}
	=
	\frac{v_{i,j+1} - v_{i,j}}{\Delta t}.
	$$

	\renewcommand{\algorithmicrequire}{\textbf{Input:}}
	\renewcommand{\algorithmicensure}{\textbf{Output:}}
	\begin{algorithm}[!t]
		\caption{The Safety Filter for RSTP Planning.}
		\label{alg:RSTP}
		\begin{algorithmic}[1]
			\REQUIRE \revfinal{$\epsilon^{\text{traj}}_{\theta}$, $N_{\text{filter}}$, $N_{\text{retry}}$, and $\mathcal{O}_{\text{all}}$.}
			\ENSURE \revfinal{The final safe planned trajectory $\tau^\star$.}
			
			\STATE $\tau^\star \gets \emptyset$,\, $k \gets 1$;
			
			\WHILE {$k \le N_{\text{retry}}$ }
			\STATE $\mathcal{B}^{\text{test}} \gets$ generate $N_{\text{filter}}$ inference trajectories \textbf{simultaneously} using $\epsilon^{\text{traj}}_{\theta}$ constrained on $\mathcal{O}_{\text{all}}$;
			
			\STATE $\mathbf{q}_{i^\star}, \boldsymbol{\varphi}_{i^\star}$  $\gets$  find the minimum cost $J_{i^\star}$ in $\mathcal{B}^{\text{test}}_{1:N_{\text{filter}}}$ and its corresponding trajectory; \% \textcolor[rgb]{0.4, 0.4, 0.4}
			{Apply safety filter to obtain the most feasible trajectory in the batch.}
			
			\STATE {\text{\textbf{if}}  $J_{i^\star} < V_{\text{inf}}$ \textbf{and} \revfinal{$\phi_{i^\star}$} $< V_{\text{inf}}$ \text{\textbf{then}} \% \textcolor{purple}{Motion limits.}} 
			
			\STATE $\quad \tau^\star \gets \mathbf{q}_{i^\star}, \boldsymbol{\varphi}_{i^\star}$;
			\STATE $\quad$ Return $\tau^\star$; \% \textcolor[rgb]{0.4, 0.4, 0.4}{ Find the optimal trajectory.}
			\STATE {\text{\textbf{end if}}}
			\STATE $k\gets k+1$;  \% \textcolor{purple}{ Replanning.}
			\ENDWHILE
			\STATE Return $\emptyset$;
		\end{algorithmic}
	\end{algorithm}
	
	The yaw rate and steering angle are computed as:
	$$
	r_{i,j}
	=
	\frac{\varphi_{i,j+1} - \varphi_{i,j}}{\Delta t},
	\qquad
	\delta_{i,j}
	=
	\arctan\!\left(\frac{l_w r_{i,j}}{v_{i,j}}\right).
	$$
	where $l_w$ is the vehicle wheelbase.
	
	To maintain consistent scaling across different metrics, batch-wise min--max normalization is applied to path length $\mathcal{L}_i$, squared acceleration \revfinal{$\|a_i\|_2^2$}, and squared steering \revfinal{$\|\delta_i\|_2^2$}. The normalized quantities are denoted as $\tilde{\mathcal{L}}_i$, $\tilde{a}_i$, and $\tilde{\delta}_i$.
	
	
	\subsubsection{Collision Risk Evaluation}
	
	For each trajectory, the minimum clearance to surrounding obstacles is computed as $\rho_i$.
	A collision risk cost is defined as:
	$$
	J_{i,\text{safe}}
	=
	\frac{1}{\rho_i + \revfinal{\varepsilon_s}}\,\mathbb{I}(\rho_i > 0)
	+
	V_{\text{inf}}\,\mathbb{I}(\rho_i = 0),
	$$
	where \revfinal{$\varepsilon_s>0$ prevents division by zero} and $V_{\text{inf}}$ is a large penalty assigned to colliding trajectories, \revfinal{set to $10^6$ in all experiments.}
	\revkeep{This formulation automatically removes unsafe candidates while selecting trajectories with larger obstacle clearance.}{This penalizes collisions and favors larger clearance.}
	
	
	\subsubsection{Cost Aggregation and Selection}
	
	The final trajectory cost used for trajectory selection is defined as:
	\begin{equation}
		\begin{aligned}
			J_i
			=
			\omega_1 \tilde{\mathcal{L}}_i
			+
			\omega_2 \|\tilde{a}_i\|_2
			+
			\omega_3 \|\tilde{\delta}_i\|_2
			+
			\omega_4 J_{i,\text{safe}},
			\quad i=1,\dots,N_{\text{filter}}.
		\end{aligned}
		\label{eq:sum_cost}
	\end{equation}
	where \revfinal{$\omega_1$--$\omega_4$} are tunable weights.
	The trajectory with the minimum cost is selected for execution:
	$$
	i^\star = \arg\min_i J_i.
	$$
	
	
	\subsubsection{Kinematic Feasibility Check}
	
	To enforce actuator limits, we additionally constrain yaw rate variation.
	Let:
	$$
	\revfinal{\Delta \boldsymbol{\varphi}_i}
	=
	\revfinal{\boldsymbol{\varphi}_{i,[2:L]} -
		\boldsymbol{\varphi}_{i,[1:L-1]}}.
	$$
	If any yaw increment exceeds the physical bound $r_{\max}\Delta t$, where $r_{\max}$ is the maximum yaw rate, a large penalty is applied:
	$$
	\revfinal{\phi_i} =
	\begin{cases}
		0, & \revfinal{|\Delta \boldsymbol{\varphi}_i| \le r_{\max}\Delta t}, \\
		V_{\text{inf}}, & \text{otherwise}.
	\end{cases}
	$$
	\revkeep{This step ensures compliance with vehicle yaw rate limits. Finally, if replanning exceeds $N_{\text{retry}}$ attempts without success, the planner returns an empty result and the vehicle stops.}{If all $N_{\text{retry}}$ trials fail this check, the planner returns $\emptyset$ and the vehicle stops.}
	
	\endgroup
	\subsection{MPC-based Data Generation with Kinematic Feasibility}\label{section:dataset_generation}
	\revshort{This section describes the dataset generation process for model learning as  shown in Fig.~\ref{fig:pipeline}. We propose a novel MPC formulation to generate kinematically feasible trajectories.
		The vehicle uses the kinematics model as in~\cite{zhang2020optimization}, with state $\boldsymbol{\zeta} = [x, y, \varphi, v]^\top$ and control input $\mathbf{u} = [\delta, a]^\top$.
		The kinematic model is discretized via the improved Euler method. The MPC optimization problem over a prediction horizon $N_p$ with \revfinal{sampling time $T_s$} is:}{For dataset generation (Fig.~\ref{fig:pipeline}), an MPC planner produces feasible trajectories with state $\boldsymbol{\zeta}=[x,y,\varphi,v]^\top$ and input $\mathbf{u}=[\delta,a]^\top$~\cite{zhang2020optimization}. With improved Euler discretization over horizon $N_p$ and sampling time $T_s$, the MPC optimization problem is formulated as follows:}
	\begin{equation}
		\begin{aligned}\label{eq:mpc}
			\min_{\boldsymbol{\zeta}, \mathbf{u}} \quad & \gamma \cdot J_\textrm{dynamic} +   \sum_{k=1}^{N_p}   \overbrace{ \left( \Vert \mathbf{A} \cdot \boldsymbol{\zeta}_k - \mathbf{p}_{s_k}^{\text{ref}} \Vert^2_{Q_1} \right)}^{J_\textrm{static}}  \\
			& + \sum_{k=1}^{N_p-1}\underbrace{ \Vert \Delta \mathbf{u}_k \Vert^2_{R_1}
				+ \Vert \mathbf{u}_k \Vert^2_{R_2}
				 + \Vert \mathbf{A} \cdot \boldsymbol{\zeta}_{N_p} - \mathbf{p}_{s_{N_p}}^{\text{ref}} \Vert^2_{Q_2}     }_{J_\textrm{static}} \\
			\text{s.t.} \quad
			& \boldsymbol{\zeta}_{k} = \boldsymbol{\zeta}_{k-1} + T_s \cdot f \left( \boldsymbol{\zeta}_{k-1} + \frac{T_s}{2} f \left( \boldsymbol{\zeta}_{k-1}, \mathbf{u}_k \right), \mathbf{u}_k \right), \\
			& \boldsymbol{\zeta}_0 = \boldsymbol{\zeta}_{\textrm{cur}}, \quad \revfinal{\boldsymbol{\zeta}_{\textrm{min}} \leq \boldsymbol{\zeta}_k \leq \boldsymbol{\zeta}_{\textrm{max}}, \quad \mathbf{u}_{\textrm{min}} \leq \mathbf{u}_k \leq \mathbf{u}_{\textrm{max}}},
		\end{aligned}
	\end{equation}
	\revkeep{where \revfinal{$\mathbf{A}=[\,\mathbf{I}_3\ \mathbf{0}_{3\times1}\,]\in\mathbb{R}^{3\times4}$ selects $(x,y,\varphi)$ from $\boldsymbol{\zeta}_k$}, and $Q_1$, $Q_2$, $R_1$, and $R_2$ are weighting matrices. \revfinal{The state and input bounds are element-wise box constraints.} $\boldsymbol{\zeta}_{\text{cur}}$ represents the current state. The scalar $\gamma$ is used to select between static and dynamic scenes. For static scene dataset generation, the MPC planner simplifies to reference tracking by setting $\gamma = 0$. For dynamic scene dataset generation, a collision penalty is introduced: $J_{\textrm{dynamic}} = \alpha \sum_{i=1}^{N_o} \|\mathbf{p}_{\textrm{cur}} - \mathbf{o}_i\|_2^{-1},$ where $N_o$ is the number of dynamic obstacles and $\mathbf{o}_i$ represents their positions. In this case, with $\gamma = 1$, static obstacles are excluded, allowing potential collisions with them, as illustrated in Fig.~\ref{fig:pipeline}. The reference pose sequence $\mathbf{p}_{s_k}^{\text{ref}} = [x_{s_k}^{\text{ref}}, y_{s_k}^{\text{ref}}, \varphi_{s_k}^{\text{ref}}]^\top$ is parameterized by the arc length $s_k$, where \revfinal{$s_k = s_{k-1} + T_s \cdot v_{\text{ref}}$}, with $s_0 = s_{\text{cur}}$.}{where $\mathbf{A}=[\,\mathbf{I}_3\ \mathbf{0}_{3\times1}\,]\in\mathbb{R}^{3\times4}$ selects $(x,y,\varphi)$ from $\boldsymbol{\zeta}_k$, $Q_1,Q_2,R_1,R_2$ are weights, and $\boldsymbol{\zeta}_{\text{cur}}$ is the current state. The scalar $\gamma$ switches between static tracking ($\gamma=0$) and dynamic-scene generation ($\gamma=1$) with collision penalty $J_{\textrm{dynamic}}=\alpha\sum_{i=1}^{N_o}\|\mathbf{p}_{\textrm{cur}}-\mathbf{o}_i\|_2^{-1}$. The reference pose $\mathbf{p}_{s_k}^{\text{ref}}=[x_{s_k}^{\text{ref}},y_{s_k}^{\text{ref}},\varphi_{s_k}^{\text{ref}}]^\top$ is indexed by arc length $s_k=s_{k-1}+T_s\,v_{\text{ref}}$, $s_0=s_{\text{cur}}$.}

	\revkeep{During the dataset generation process, reference paths are planned using the method proposed in ~\cite{liRapidIterativeTrajectory}. \revmwl{While these serve as coarse guides on the known static map, dynamic obstacles are introduced through $J_{\textrm{dynamic}}$.} The initial state is randomly sampled within the feasible region, while the goal state is fixed with zero velocity.  To maintain a consistent trajectory horizon $L$, trajectories shorter than $L$ are padded by duplicating the last value, while longer trajectories are truncated. Each demonstration trajectory is represented as $\boldsymbol{\tau}_i^{\text{dem}} \in \mathbb{R}^{4 \times L}$, and in dynamic scenes the corresponding obstacle trajectory is denoted as $\mathcal{O}_i^{\text{single}} \in \mathbb{R}^{2 \times L}$.}{Reference paths from~\cite{liRapidIterativeTrajectory} provide coarse static-map guides, while dynamic obstacles enter through $J_{\textrm{dynamic}}$. Initial states are sampled from feasible regions, goals have zero velocity, and trajectories are padded or truncated to horizon $L$, yielding $\boldsymbol{\tau}_i^{\text{dem}}\in\mathbb{R}^{4\times L}$ and, for dynamic scenes, $\mathcal{O}_i^{\text{single}}\in\mathbb{R}^{2\times L}$.}
	
	
	\section{EXPERIMENTAL RESULTS AND DISCUSSION} \label{sec:Exp}
	
	\revkeep{The proposed RSTP method is validated through simulation and real-world experiments. Simulations evaluate safety, efficiency, and kinematic feasibility, and confirm the safety filter's effectiveness. Experiments on the F1TENTH platform further demonstrate robustness under real-world uncertainties.}{Simulations evaluate safety, efficiency, and kinematic feasibility, while F1TENTH tests assess real-world robustness.}
	
	\subsection{Experiment Settings}\label{app:sub1}
	\revkeep{To evaluate RSTP, training and testing datasets are generated from a real-world $6\text{m} \times 6\text{m}$ map. A test set $D_A$ is constructed by uniformly sampling 1045 initial poses (\revfinal{$N_{D,\mathrm{all}} = 1045$}) from the navigable area. From this set, a subset $D_C$ containing 79 poses (\revfinal{$N_{D,\mathrm{comp}} = 79$}) is selected to evaluate compositional performance. Target poses are fixed. The trajectory horizon is $L=128$, with 100 diffusion steps during training and 8 DDIM steps for denoising. The safety filter samples $N_{\text{filter}}=8$ candidate trajectories. Experiments are conducted on an RTX 4060 GPU and an Intel i7-12700 CPU (4.9 GHz, 48 GB RAM). The diffusion model is built upon a 1D U-Net backbone with channel dimensions of 16, 64, and 128. It is trained using Adam with a learning rate of $2 \times 10^{-4}$. Residual temporal blocks with 1D convolutions are used. The training and inference pipeline of the diffusion model follows the architecture outlined in \cite{luo2024potential}, with our proposed composition formula defined in Eq.~\eqref{eq:compose}.}{Datasets are generated on a real $6\text{m}\times6\text{m}$ map. The test set $D_A$ contains 1045 initial poses, and $D_C$ contains 79 poses for composition evaluation. Targets are fixed, $L=128$, training uses 100 diffusion steps, denoising uses 8 DDIM steps, and the filter samples $N_{\text{filter}}=8$ candidates. Experiments run on an RTX 4060 GPU and Intel i7-12700 CPU. The 1D U-Net follows~\cite{luo2024potential} with channels 16/64/128 and Adam learning rate $2\times10^{-4}$.}

	\subsection{Environments and Baselines}\label{sec:Exp_setup}
\begin{figure*}[!t]
		\centering
		\includegraphics[width=6.85in]{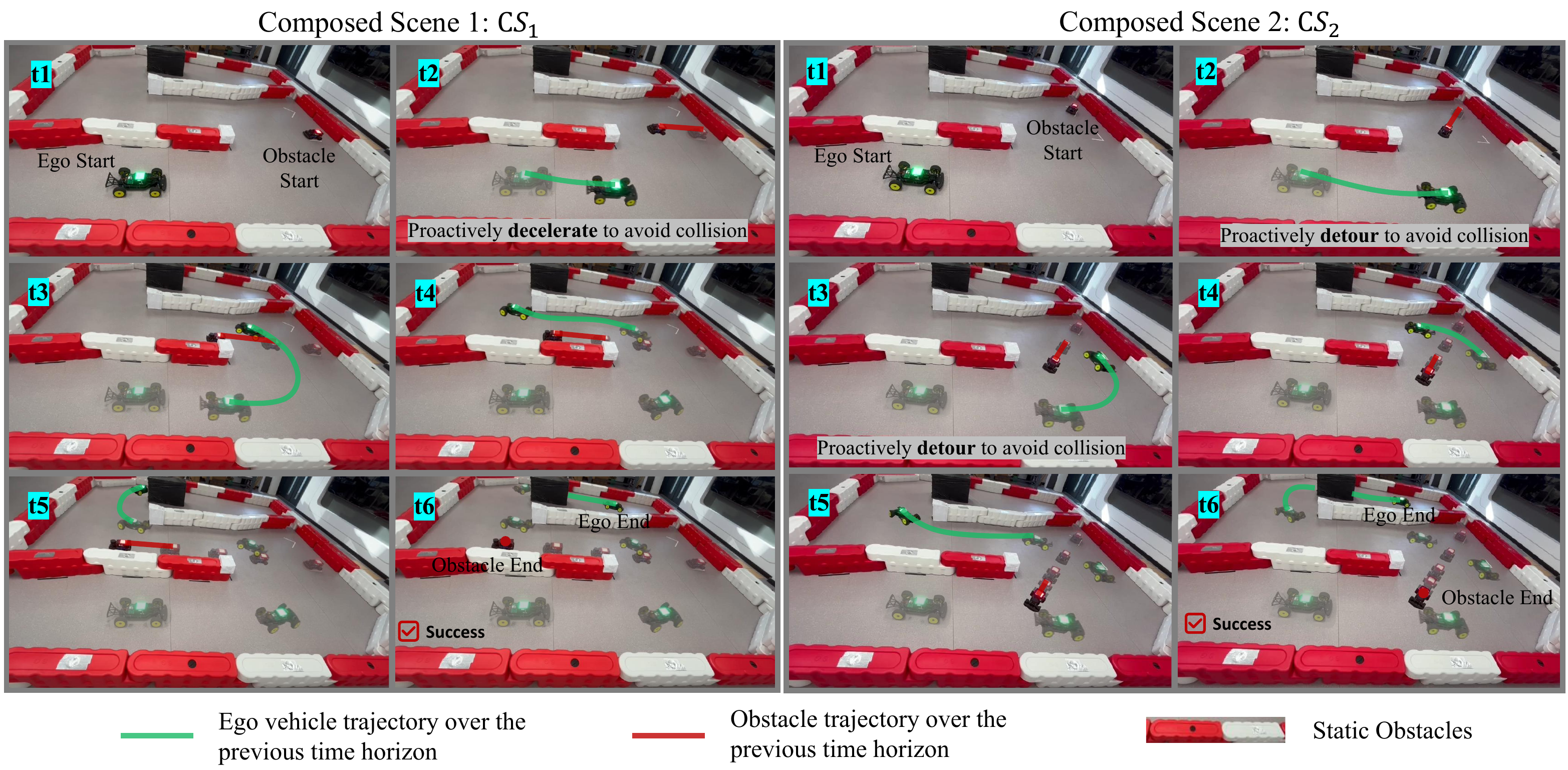}
		\caption{\revshort{\textbf{Test-Time Diffusion Composition Enables Safe Generalization in Unseen Dynamic Environments.}
				In Composed Scene 1, the ego vehicle proactively decelerates when a dynamic obstacle enters a potential conflict region, thereby avoiding imminent collision. In Composed Scene 2, it instead executes a smooth detour maneuver to safely bypass the obstacle while maintaining trajectory stability.}{\textbf{Real-world generalization in unseen dynamic scenes.} The ego vehicle decelerates in $CS_1$ and detours in $CS_2$ to avoid the moving obstacle.}}
		\label{fig:exp_all}
	\end{figure*}

	\revshort{In all experiments, equal weights are assigned to the terms in Eq.~\eqref{eq:sum_cost}, and start/goal states are enforced via inpainting.
		We evaluate three scenarios (as \revfinal{shown} in Fig.~\ref{fig:exp_all}):}{Eq.~\eqref{eq:sum_cost} uses equal weights, and start/goal states are enforced by inpainting. Three scenarios are evaluated (Fig.~\ref{fig:exp_all}):}
	\begin{itemize}
		\item
		Static Scene (SS): Contains only static obstacles. The corresponding trained model is denoted as Static Model (StM).
		\item
		Composed Scene 1 ($CS_1$): SS with one moving vehicle initialized at $[3.5, 1.2, 1.0, 0.00]$, moving straight at $0.4\,\text{m/s}$ for $7.1\,\text{s}$ (left in Fig.~\ref{fig:exp_all}). The model trained on this trajectory is denoted as DyM1.
		
		\item
		Composed Scene 2 ($CS_2$): SS with one moving vehicle initialized at $[3.5, 2.0, 0.891, -0.454]$, moving straight at $0.4\,\text{m/s}$ for $5.3\,\text{s}$ (right in Fig.~\ref{fig:exp_all}). The corresponding model is DyM2.

	\end{itemize}
	
	\revkeep{We compare RSTP with search-based methods $\text{A}^\star$~\cite{liu2017global}, $\text{RRT}^\star$~\cite{karaman2011sampling}, and $\text{FMT}^\star$~\cite{wu2021st}, which ignore vehicle kinematics, as well as kinematics-aware planners $\text{Kino RRT}$~\cite{sakcak2019sampling} and $\text{Hybrid A}^\star$~\cite{dolgov2008practical} (Baseline, denoted as BLM). \revfinal{We also compare with optimization-based methods, including RITP~\cite{liRapidIterativeTrajectory} and OBCA~\cite{zhang2020optimization}.} We additionally include a conventional diffuser~\cite{ho2020denoising} without energy modeling and the MPPI~\cite{honda2025model} control method as baselines for comparison. Kinematic feasibility is evaluated using the MPC tracker (Section~\ref{sec:Method}) with Mean Tracking Error (M.TE), while planning efficiency is measured by Computation Time (C.T). For RSTP, C.T includes replanning time. The percentage reduction relative to BLM is defined as \revfinal{$\text{M.RP}=\frac{\text{C.T}_{\text{BLM}}-\text{C.T}}{\text{C.T}_{\text{BLM}}}$} for time and \revfinal{$\text{M.RP}=\frac{\text{M.TE}_{\text{BLM}}-\text{M.TE}}{\text{M.TE}_{\text{BLM}}}$} for tracking error. RSTP without the safety filter is denoted as RSTP-NSF. OBCA and RITP are warm-started by $\text{Hybrid A}^\star$, and MPPI by $\text{A}^\star$.}{Baselines include $\text{A}^\star$~\cite{liu2017global}, $\text{RRT}^\star$~\cite{karaman2011sampling}, $\text{FMT}^\star$~\cite{wu2021st}, Kino RRT~\cite{sakcak2019sampling}, Hybrid $\text{A}^\star$ (BLM)~\cite{dolgov2008practical}, RITP~\cite{liRapidIterativeTrajectory}, OBCA~\cite{zhang2020optimization}, a non-energy diffuser~\cite{ho2020denoising}, and MPPI~\cite{honda2025model}. We report computation time (C.T), mean tracking error (M.TE), and relative reduction $\text{M.RP}=(\text{BLM}-\text{method})/\text{BLM}$. RSTP-NSF disables the safety filter; RITP/OBCA use Hybrid $\text{A}^\star$ warm starts and MPPI uses $\text{A}^\star$.}
	\vspace{-10pt}

	\begin{table*}[!t]
		\centering
		\caption{\revkeep{Simulation Results on the Static Scene: Validation of Time Efficiency and Safety of the RSTP Method. Warm Start (WS) paths are planned by Hybrid A* method.}{Simulation Results in the Static Scene. WS denotes Hybrid A* warm start.}}
		\footnotesize
		\setlength{\tabcolsep}{2.0pt}
		\renewcommand{\arraystretch}{1.05}
		\begin{tabular*}{\textwidth}{@{\extracolsep{\fill}}c|c|cc|cccccc|ccccc@{}}
			\toprule
			\multicolumn{2}{c|}{\multirow{2}[4]{*}{Methods}} & \multirow{2}[4]{*}{F.Rate$^1$} & \multirow{2}[4]{*}{C.Rate$^2$} & \multicolumn{6}{c|}{Computation Time (s)}              & \multicolumn{5}{c}{Tracking Error} \\
			\cmidrule{5-15}    \multicolumn{2}{c|}{} &       &       & Max   & Min   & Mean  & M.RP  & Var & WS   & Max   & Min   & Mean  & M.RP  & Var \\
			\midrule
			\midrule
			
			\multirow{5}[2]{*}{Heuristics} 
			& $\text{FMT}^\star$  & 0.77\% & 74.07\% & 2.766 & 0.615 & 1.450 & 92.27\% & 0.124 & / & 10.194 & 0.284 & 5.188 & -92.23\% & 4.237 \\
			& $\text{A}^\star$    & 0.86\% & 67.37\% & 5.337 & 0.138 & 2.032 & 89.16\% & 0.924 & / &  53.129 & 1.085 & 3.903 & -44.62\% & 8.772 \\
			& $\text{RRT}^\star$  & 1.53\% & 59.43\% &  13.950 & 0.160 & 3.177 & 83.05\% & 3.029 & / & 23.561 & 0.951 & 4.290 & -58.96\% & 12.235 \\
			& $\text{Kino RRT}$ & 14.64\% & \multicolumn{1}{c|}{58.18\%} & 39.362 & 0.019 & 4.267 & 77.24\% & {31.601} & / & 39.686 & 0.476 & 3.232 & -19.77\% & 16.876 \\
			& $\text{Hybrid A}^\star$ & 24.88\% & 27.75\% & 248.964 & {0.004} & 18.747 & -     & $\uparrow$ & / & 6.929 & 0.459 & 2.699 & -     & 0.882 \\
			
			\midrule
			\multirow{2}[2]{*}{Optimization} 
			
			& RITP  & {0.76\%} & {7.87\%} & {0.351} & {0.008} & {0.059} & {99.69\%} & 0.002 & 18.7 & {6.413} & {0.254} & {0.792} & {70.67\%} & {0.377} \\
			& OBCA  & 10.41\% & 13.20\% & 1.395 & 0.378 & 0.508 & 97.29\% & 0.005 & 18.7 & {5.877} & {0.178} & {0.560} & {79.27\%} & {0.111} \\
			
			\midrule
			
			\multirow{4}[2]{*}{Sampling} 
			
			& MPPI  & 3.16\% & 82.97\% & 0.109 & 0.010 & 0.023 & 99.88\% & 3.25e-4 & 2.0 & 49.655 & 0.871 & 5.935 & -119.91\% & 55.584 \\
			& Diffuser  & 9.13\% & 17.51\% & {0.259} & 0.043 & {0.044} & {99.77\%} & {4.98e-5} & / & 46.193 & 0.309 & 1.380 & 48.89\% & 7.912 \\
			
			& RSTP-NSF & 4.50\% & 11.87\% & 0.515 & 0.144 & 0.195 & 98.96\% & {1.65e-4} & / & 37.933 & 0.284 & 1.495 & 44.62\% & 4.299 \\
			
			& RSTP (ours)  & \cellcolor[rgb]{ .949,  .949,  .949} {\textbf{\textcolor[rgb]{ .753,  0,  0}{0.57\%}}} & \cellcolor[rgb]{ .949,  .949,  .949} {\textbf{8.80\%}} & \cellcolor[rgb]{ .949,  .949,  .949} 0.522 & \cellcolor[rgb]{ .949,  .949,  .949} 0.137 & \cellcolor[rgb]{ .949,  .949,  .949} {\textcolor[rgb]{ .753,  0,  0}{\textbf{0.210}}} & \cellcolor[rgb]{ .949,  .949,  .949}98.88\% & \cellcolor[rgb]{ .949,  .949,  .949} {{3.16e-4}} & / & \cellcolor[rgb]{ .949,  .949,  .949} 12.909 & \cellcolor[rgb]{ .949,  .949,  .949} 0.321 & \cellcolor[rgb]{ .949,  .949,  .949} {{1.424}} & \cellcolor[rgb]{ .949,  .949,  .949} 47.23\% & \cellcolor[rgb]{ .949,  .949,  .949} 1.647 \\
			\bottomrule
		\end{tabular*}%
		
		\begin{tablenotes}
			\footnotesize
			\item$^1$ \revkeep{\revfinal{F.Rate=$N_f/N_{D,\mathrm{all}}$}, where $N_{D,\mathrm{all}}$ is the number of test cases and $N_f$ denotes planning failures. A planning failure means that no geometrically collision-free candidate is generated within the specified iterations and time.}{F.Rate=$N_f/N_{D,\mathrm{all}}$, where $N_f$ is the number of planning failures.}
			\item$^2$ \revkeep{\revfinal{C.Rate=$(N_f+N_c)/N_{D,\mathrm{all}}$}, where $N_c$ counts trajectories that collide during closed-loop tracking.}{C.Rate=$(N_f+N_c)/N_{D,\mathrm{all}}$, where $N_c$ counts tracking collisions.}
		\end{tablenotes}
		\label{tab:main_exp}%
		\vspace{-5pt}
	\end{table*}%

	\subsection{Evaluate the Performance of RSTP Method in Simulation}
	
	\revkeep{We first evaluate the \revfinal{safety-related metrics} and efficiency of RSTP, then validate the safety filter, and finally assess how the composition of energy-parameterized diffusion models generalizes to new scenes. Experimental results are summarized in Table~\ref{tab:main_exp}.}{Table~\ref{tab:main_exp} summarizes safety, efficiency, filter ablation, and composition.}

	\subsubsection{On the Safety Performance of our RSTP method}
\begin{figure*}[!t]
		\centering
		\includegraphics[width=7.0in]{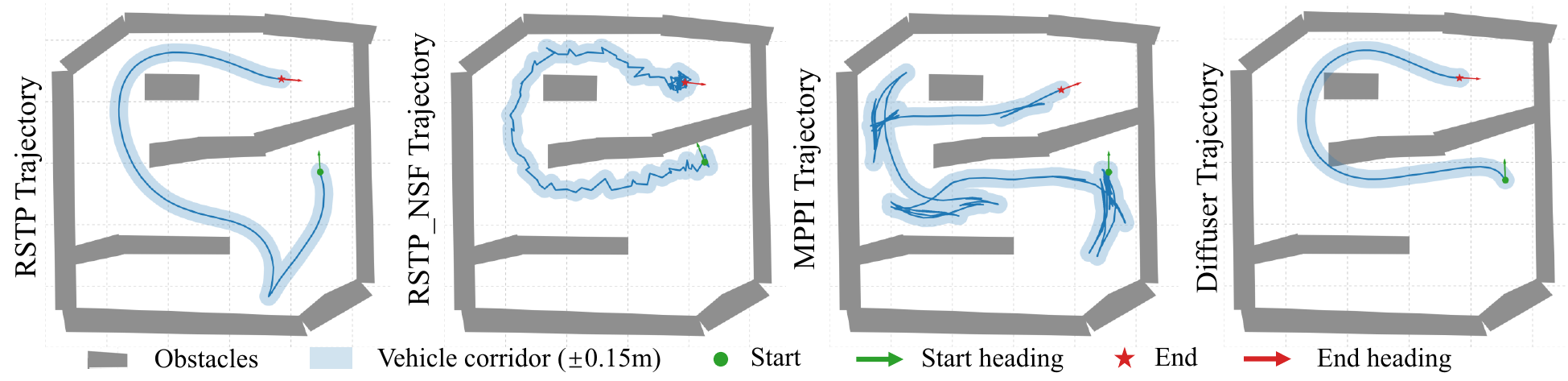}
		\caption{\revkeep{\textbf{The Proposed RSTP Method Generates Smooth, Safe, and Kinematically Feasible Trajectories in Static Scene.} In contrast, RSTP without the safety filter (RSTP-NSF) and the baseline diffuser lack explicit safety and feasibility constraints, leading to collisions.
				Although MPPI can produce collision-free and feasible solutions, its trajectories exhibit higher variance and less structured behavior, potentially compromising tracking stability.}{\textbf{Static-scene planning comparison.} RSTP generates a smooth, safe, and feasible trajectory; RSTP-NSF and diffuser collide, while MPPI shows higher trajectory variance.}}
		\label{fig:sim_env}
		\vspace{-5pt}
	\end{figure*}

	\revshort{\revmwl{Fig.~\ref{fig:sim_env} compares four sampling-based planners under the same start and goal states. RSTP generates a smooth, collision-free trajectory with necessary gear shifts. Without trajectory screening, RSTP-NSF diverges during planning, indicating that raw diffusion candidates can be unstable. The conventional diffuser lacks explicit safety and feasibility constraints and produces a collision. MPPI can occasionally remain collision-free, but its unimodal sampling stays near the current control mode, causing frequent gear switches and missing long-horizon maneuvers such as the initial reverse motion.}}{As shown in Fig.~\ref{fig:sim_env}, among the sampling-based methods, only RSTP generates a smooth, collision-free trajectory with necessary gear shifts. In contrast, RSTP-NSF diverges during planning, suggesting that raw diffusion samples are not always reliable without test-time screening. The conventional diffuser lacks explicit safety and feasibility constraints and produces a collision. MPPI can occasionally remain collision-free, but its unimodal sampling stays close to the current control mode, which makes the planner sensitive to local optima and leads to frequent gear switches. This also explains why it misses long-horizon maneuvers such as the initial reverse motion.}

	\revshort{The quantitative results summarized in Table~\ref{tab:main_exp} show that the proposed RSTP achieves a Failure Rate (F.Rate) of only \textbf{0.57\%} and a low Collision Rate (C.Rate) of {8.8\%} during tracking. It shows RSTP significantly reduces collisions during tracking,  outperforming kinematics-aware methods such as Hybrid $\text{A}^\star$ and Kino $\text{RRT}$, which suffer from higher failure rates of 24.88\% and 14.64\% due to the limitations of discrete search and sampling in long-horizon planning. Optimization-based methods, RITP and OBCA, explicitly enforce kinematics and achieve comparable collision rates of 7.87\% and 13.2\%, compared to RSTP. However, they require well-defined warm-start paths, which increase computation time.  In summary, RSTP delivers notable safety and efficiency gains, benefits from the energy-parameterized diffusion model's generative ability, and stability.}{The quantitative results in Table~\ref{tab:main_exp} show that RSTP achieves a Failure Rate (F.Rate) of only \textbf{0.57\%} and a Collision Rate (C.Rate) of \textbf{8.8\%} during tracking. \revfinal{F.Rate measures planning failure, whereas C.Rate includes planning failures and closed-loop tracking collisions.} RSTP reduces collisions compared with kinematics-aware methods such as Hybrid $\text{A}^\star$ and Kino $\text{RRT}$, whose higher failure rates of 24.88\% and 14.64\% reflect the limitations of discrete search and sampling in long-horizon planning. Optimization-based methods, RITP and OBCA, explicitly enforce kinematics and achieve comparable collision rates of 7.87\% and 13.2\%, but they require well-defined warm-start paths and thus increase computation time. Overall, RSTP improves planning success and tracking safety while preserving efficiency.}

	\subsubsection{On the Time Efficiency of our RSTP method}
	
	\revshort{As shown in Table~\ref{tab:main_exp}, the proposed $\text{RSTP}$ achieves a mean C.T of only \textbf{0.21s} across all initial poses in $D_A$, demonstrating strong real-time performance. It also exhibits high stability, with a variance of \textbf{3.16e-4}. This efficiency benefits from state-based denoising, which significantly reduces the noise dimension.
		Although optimization-based methods (RITP and OBCA) are also fast, they rely on Hybrid $\text{A}^\star$ for initialization, introducing an additional mean cost of 18.7s. MPPI and the diffuser are computationally efficient as well, but their trajectory quality is noticeably worse than that of RSTP, as illustrated in Fig.~\ref{fig:sim_env}. \revmwl{For MPPI, the low mean C.T reflects short-horizon sampling rather than reliable global planning, as confirmed by its high C.Rate and M.TE.}}{As shown in Table~\ref{tab:main_exp}, the proposed $\text{RSTP}$ achieves a mean C.T of only \textbf{0.21s} across all initial poses in $D_A$, demonstrating strong real-time performance. It also exhibits high stability, with a variance of \textbf{3.16e-4}. This efficiency benefits from state-based denoising, which significantly reduces the noise dimension. Although optimization-based methods (RITP and OBCA) are also fast, they rely on Hybrid $\text{A}^\star$ for initialization, introducing an additional mean cost of 18.7s. MPPI and the diffuser are computationally efficient as well, but their trajectory quality is noticeably worse than that of RSTP, as illustrated in Fig.~\ref{fig:sim_env}. }
	\vspace{-5pt}

	\subsubsection{On the Effectiveness of Safety Filter}
	\revshort{As shown in Fig.~\ref{fig:sim_env}, the safety filter can efficiently select kinematically feasible and safe trajectories. However, without the safety filter, the F.Rate and C.Rate increase to \textbf{4.50\%} and \textbf{11.87\%}, respectively, highlighting the necessity of the safety filter. The mean C.T of RSTP-NSF is 0.195 s, comparable to 0.21 s with the filter enabled, indicating negligible computational overhead and preserving time efficiency. \revmwl{The slightly smaller minimum C.T of RSTP does not indicate that filtering accelerates planning; the Min column is a single best-case timing, not a paired comparison between RSTP and RSTP-NSF on the same sample, and is sensitive to runtime fluctuation. The mean C.T better reflects the small additional filtering cost. Failures mainly occur when denoising yields a low-diversity batch near an unsafe mode; the filter can reject candidates but cannot create a new feasible one.}}{As shown in Fig.~\ref{fig:sim_env}, the safety filter efficiently selects \revfinal{candidates satisfying obstacle clearance and kinematic checks}. Without it, the F.Rate and C.Rate rise to \textbf{4.50\%} and \textbf{11.87\%}. The mean C.T of RSTP-NSF is 0.195 s, comparable to 0.21 s with the filter enabled, so the filter adds negligible overhead while preserving time efficiency. The slightly smaller minimum C.T of RSTP does not indicate that filtering accelerates planning; the Min column is a best-case timing, not a paired comparison on the same sample, and is sensitive to runtime fluctuation. The mean C.T better reflects the small additional filtering cost.}
	
	\subsection{RSTP Performance in Real-world F1TENTH Vehicle}

	\begin{table}[!t]
		\centering
		\caption{\revshort{The Vehicle--Obstacle Distances in Real-world Execution when Tracking the RSTP Trajectory.}{Vehicle--Obstacle Distances during Real-World Tracking.}}
		\begin{adjustbox}{max width=\columnwidth}
			\footnotesize
			\setlength{\tabcolsep}{5pt}
			\renewcommand{\arraystretch}{1.05}
			\begin{tabular}{c|c|ccccc}
				\toprule
				Scenes                     & Start points   & Danger$^1$                                     & Max   & Min   & Mean  & Std   \\
				\midrule
				\midrule
				\multirow{3}[1]{*}{$CS_1$} & [0.86, 0.03]   & 6.60\%                                          & 0.591 & {{0.032}} & 0.283 & 0.020 \\
				& [3.03, -0.07]  & 8.54\%                                          & 0.559 & 0.038 & \textcolor[rgb]{ .753,  0,  0} {\textbf{0.260}} & 0.020 \\
				& [2.73, 0.32]   & 3.77\%                                          & 0.532 & 0.054 & 0.288 & 0.016 \\
				\midrule
				\multirow{3}[1]{*}{$CS_2$} & [0.82, 0.07]   & 4.55\%   & 0.590 & 0.051 & 0.278 & 0.015 \\
				& [2.76, 0.47]   & \textcolor[rgb]{ .753,  0,  0}{\textbf{0.00\%}} & 0.593 & 0.086 & 0.304 & 0.019 \\
				& [2.79, -0.11]  & 7.41\%                                          & 0.550 & 0.036 & 0.275 & 0.018 \\
				\bottomrule
			\end{tabular}%
		\end{adjustbox}
		
		\begin{tablenotes}
			\footnotesize
			\item $^1$ \revshort{Denoted as $\frac{N_{\text{Dgr}}}{N_{all}}$, where $N_{\text{Dgr}}$ counts points within 0.08 m of obstacles.}{$N_{\text{Dgr}}/N_{all}$; $N_{\text{Dgr}}$ counts points within 0.08 m of obstacles.}
		\end{tablenotes}
		\label{tab:real_comp}%
		\vspace{-5pt}
	\end{table}%
	
	\revkeep{The real-world experiments are conducted using a self-built F1TENTH vehicle. For mapping and localization, the vehicle utilizes the Cartographer ROS framework~\cite{hess2016real}. Communication between the server and onboard computer is managed via the Message Queuing Telemetry Transport (MQTT)  protocol. The F1TENTH vehicle uses the pure pursuit controller proposed in~\cite{weng2024aggressive} to track trajectories.}{Real-world tests use a self-built F1TENTH vehicle with Cartographer localization~\cite{hess2016real}, MQTT communication, and pure-pursuit tracking~\cite{weng2024aggressive}.}

	\revkeep{As shown in Fig.~\ref{fig:exp_all}, the proposed RSTP method \revfinal{adapts to} previously unseen dynamic environments on the real-world F1TENTH platform. \revfinal{This shows that} by composing energy-parameterized diffusion models at test time, the vehicle can adapt to new scenes without additional retraining. In Composed Scene 1 ($CS_1$), the ego vehicle proactively decelerates to avoid a potential collision when the dynamic obstacle enters a conflict region. In Composed Scene 2 ($CS_2$), the ego vehicle instead performs a detour maneuver to bypass the obstacle. These adaptive behaviors demonstrate that test-time diffusion composition enables stable trajectory generation under out-of-distribution dynamic scenarios.}{Fig.~\ref{fig:exp_all} shows that test-time composition generalizes to unseen dynamic scenes without retraining: the vehicle decelerates in $CS_1$ and detours in $CS_2$.}

	\begin{figure}[!t]
		\centering
		\includegraphics[scale=0.40]{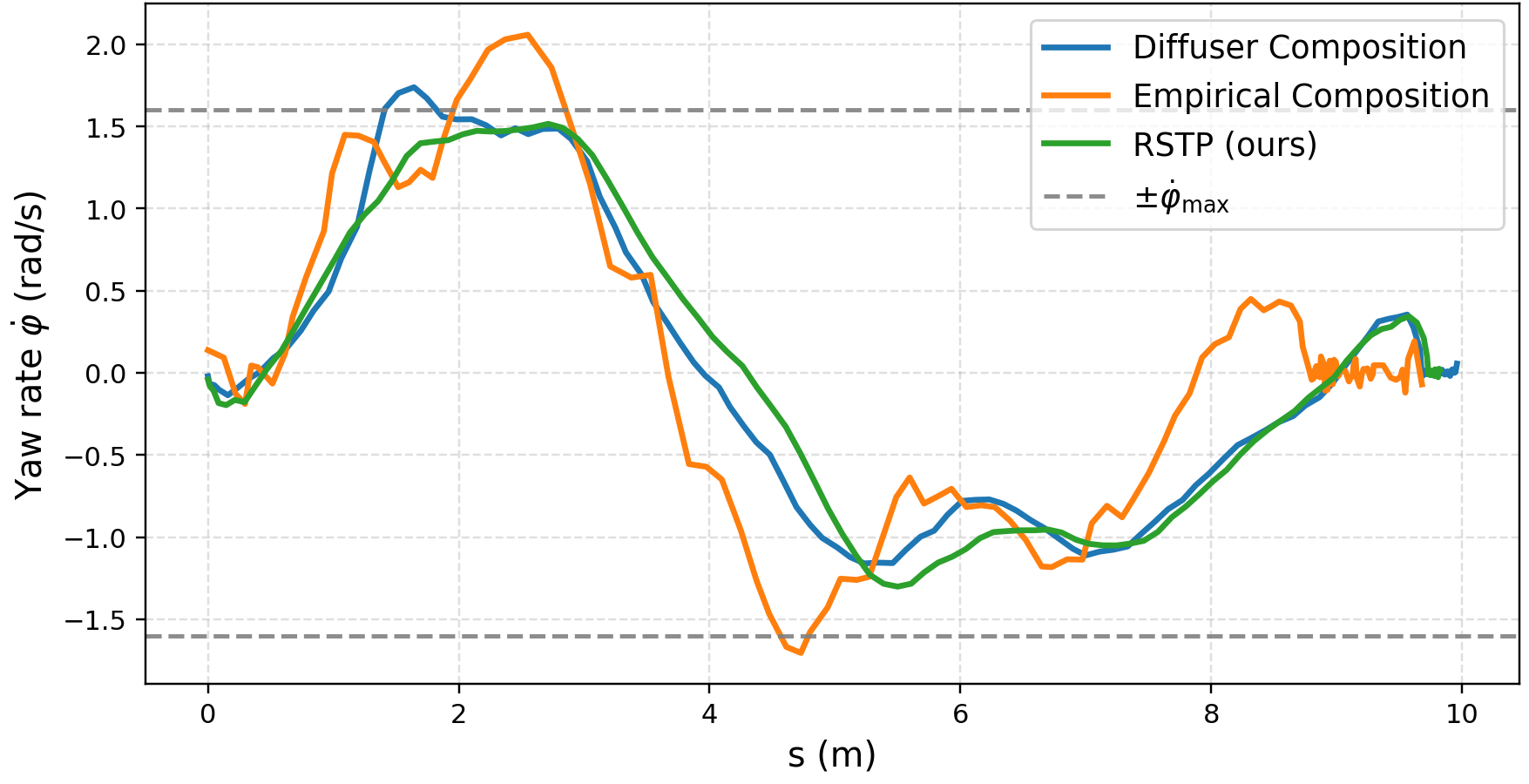}
		\caption{\revkeep{\textbf{RSTP Ensures Kinematically Feasible, Consistent Trajectory Composition.} Compared strategies produce excessive yaw rate.}{\textbf{Yaw-rate feasibility.} Baselines produce excessive yaw rates, while RSTP remains within limits.}}
		\label{fig:yaw_rate}
		\vspace{-5pt}
	\end{figure}
	
	\revshort{To further validate the proposed RSTP framework, we compare it with two alternative composition strategies: (1) the empirical formula in~\cite{luo2024potential}, and (2) a naive diffuser-based composition without energy parameterization, implemented using our unified composition scheme. We evaluate kinematic feasibility by analyzing the yaw rate profiles along the planned trajectories, \revmwl{as yaw rate indicates steering smoothness and dynamic consistency, but not an explicit denoiser constraint} (see Fig.~\ref{fig:yaw_rate}). The results show that the method in~\cite{luo2024potential} fails to maintain multi-constraint consistency when kinematic constraints are introduced, producing trajectories with excessive yaw rate fluctuations that violate vehicle limits. The naive diffuser, lacking an energy-based formulation, also leads to unstable and dynamically infeasible behaviors. In contrast, RSTP generates smooth and kinematically consistent trajectories. This improvement is attributed to its additive energy-based diffusion composition, which preserves  consistency during test-time aggregation, together with the safety filter that enforces motion constraints. \revmwl{Thus, kinematic consistency is encouraged jointly by MPC demonstrations, energy-based composition, and safety filtering.}}{We compare RSTP with the empirical rule in~\cite{luo2024potential} and a naive diffuser composition. Yaw-rate profiles in Fig.~\ref{fig:yaw_rate} show that both baselines violate limits, whereas RSTP remains smooth and feasible. This consistency \revfinal{comes from} energy-based composition and safety filtering.}
	
	\revshort{To further evaluate the safety of the proposed RSTP method, three distinct starting points are selected within obstacle-free regions of the map in both scenes $CS_1$ and $CS_2$, with results summarized in Table \ref{tab:real_comp}. The safety metric used is the minimum distance between the vehicle vertices and both static and dynamic obstacles. Even with localization uncertainty, the mean obstacle distance recorded is \textbf{0.260 m}, confirming adequate safety margins. Moreover, fewer than 10\% of trajectory points fall within the defined danger threshold, indicating that only a minor segment of the planned trajectory poses a potential collision risk. Thus, the composed model proves to be sufficiently safe in the context of real-world uncertainty.}{Table~\ref{tab:real_comp} uses three starts in each scene and reports minimum vehicle--obstacle distance. Despite localization uncertainty, the mean clearance is \textbf{0.260 m}, and fewer than 10\% of points enter the danger threshold.}
	
	\subsection{Sensitivity Analysis of Compositional Weights}\label{app:comp_sim}
	
	\revkeep{The generalization ability of energy-based diffusion model composition is evaluated in scenes $CS_1$ and $CS_2$ (Fig.~\ref{fig:exp_all}) under four weight settings (Table~\ref{tab:comp_gen}). With appropriate weights, the composed model achieves \textbf{0\%} F.Rate and safely generalizes to unseen scenes. Increasing the weights of conditional models raises the F.Rate in both scenes, indicating that the compositional parameters can change overall performance.}{Table~\ref{tab:comp_gen} evaluates composition weights in $CS_1$ and $CS_2$. Proper weights achieve \textbf{0\%} F.Rate, whereas overly large conditional weights degrade performance.}

	\section{Conclusion} \label{sec:Conclusion}
	
	\revkeep{This paper presents RSTP, an energy-parameterized diffusion planner with a test-time safety filter. The method combines stable diffusion composition, real-time planning on vehicle states, and an MPC-based data generation pipeline, and is validated on a real-world F1TENTH platform. Experimental results demonstrate effective generalization under test-time composition. Future work will add MPC-based refinement and extend RSTP to multi-modal planning.}{This paper presents RSTP, an energy-parameterized diffusion planner with test-time safety filtering. RSTP combines MPC demonstrations, state-based denoising, and feasible trajectory selection, achieving safe generalization on an F1TENTH platform.}

	\bibliographystyle{IEEEtran}
	\bibliography{bibref}

@inproceedings{tan2025flow,
    title={Flow Matching-Based Autonomous Driving Planning with Advanced Interactive Behavior Modeling},
    author={Tianyi Tan and Yinan Zheng and Ruiming Liang and Zexu Wang and Kexin Zheng and Jinliang Zheng and Jianxiong Li and Xianyuan Zhan and Jingjing Liu},
    booktitle={The Thirty-ninth Annual Conference on Neural Information Processing Systems},
    year={2025}
}

@inproceedings{zheng2025diffusionbased,
    title={Diffusion-Based Planning for Autonomous Driving with Flexible Guidance},
    author={Yinan Zheng and Ruiming Liang and Kexin ZHENG and Jinliang Zheng and Liyuan Mao and Jianxiong Li and Weihao Gu and Rui Ai and Shengbo Eben Li and Xianyuan Zhan and Jingjing Liu},
    booktitle={The Thirteenth International Conference on Learning Representations},
    year={2025},
    url={https://openreview.net/forum?id=wM2sfVgMDH}
}

@article{0278364914528132,
    author = {John Schulman and Yan Duan and Jonathan Ho and Alex Lee and Ibrahim Awwal and Henry Bradlow and Jia Pan and Sachin Patil and Ken Goldberg and Pieter Abbeel},
    title ={Motion planning with sequential convex optimization and convex collision checking},
    journal = {The International Journal of Robotics Research},
    volume = {33},
    number = {9},
    pages = {1251-1270},
    year = {2014},
    doi = {10.1177/0278364914528132},
}

@INPROCEEDINGS{7402333,
  author={Zhijie Zhu and Schmerling, Edward and Pavone, Marco},
  booktitle={2015 54th IEEE Conference on Decision and Control (CDC)}, 
  title={A convex optimization approach to smooth trajectories for motion planning with car-like robots}, 
  year={2015},
  volume={},
  number={},
  pages={835-842},
  doi={10.1109/CDC.2015.7402333}
}

@article{liRapidIterativeTrajectory,
  title={A rapid iterative trajectory planning method for automated parking through differential flatness},
  author={Li, Zhouheng and Xie, Lei and Hu, Cheng and Su, Hongye},
  journal={Robotics and Autonomous Systems},
  volume={182},
  pages={104816},
  year={2024},
  publisher={Elsevier}
}

@article{christopher2024constrained,
  title={Constrained synthesis with projected diffusion models},
  author={Christopher, Jacob K and Baek, Stephen and Fioretto, Nando},
  journal={Advances in Neural Information Processing Systems},
  volume={37},
  pages={89307--89333},
  year={2024}
}

@inproceedings{mizuta2024cobl,
  title={Cobl-diffusion: Diffusion-based conditional robot planning in dynamic environments using control barrier and lyapunov functions},
  author={Mizuta, Kazuki and Leung, Karen},
  booktitle={2024 IROS},
  pages={13801--13808},
  year={2024},
  organization={IEEE}
}

@inproceedings{chen2014stochastic,
  title={Stochastic gradient hamiltonian monte carlo},
  author={Chen, Tianqi and Fox, Emily and Guestrin, Carlos},
  booktitle={International conference on machine learning},
  pages={1683--1691},
  year={2014},
  organization={PMLR}
}

@inproceedings{salimans2021should,
  title={Should EBMs model the energy or the score?},
  author={Salimans, Tim and Ho, Jonathan},
  booktitle={Energy based models workshop-ICLR 2021},
  year={2021}
}

@article{honda2025model,
  title={Model Predictive Control via Probabilistic Inference: A Tutorial},
  author={Honda, Kohei},
  journal={arXiv preprint arXiv:2511.08019},
  year={2025}
}

@misc{torchvision2016,
    title        = {TorchVision: PyTorch's Computer Vision library},
    author       = {TorchVision maintainers and contributors},
    year         = 2016,
    journal      = {GitHub repository},
    publisher    = {GitHub},
    howpublished = {\url{https://github.com/pytorch/vision}}
}

@inproceedings{honda2024stein,
  title={Stein variational guided model predictive path integral control: Proposal and experiments with fast maneuvering vehicles},
  author={Honda, Kohei and Akai, Naoki and Suzuki, Kosuke and Aoki, Mizuho and Hosogaya, Hirotaka and Okuda, Hiroyuki and Suzuki, Tatsuya},
  booktitle={2024 IEEE International Conference on Robotics and Automation (ICRA)},
  pages={7020--7026},
  year={2024},
  organization={IEEE}
}

@article{dolgov2008practical,
  title={Practical search techniques in path planning for autonomous driving},
  author={Dolgov, Dmitri and Thrun, Sebastian and Montemerlo, Michael and Diebel, James},
  journal={ann arbor},
  volume={1001},
  number={48105},
  pages={18--80},
  year={2008}
}

@article{li2024data,
  title={A Data-Driven Aggressive Autonomous Racing Framework Utilizing Local Trajectory Planning with Velocity Prediction},
  author={Li, Zhouheng and Zhou, Bei and Hu, Cheng and Xie, Lei and Su, Hongye},
  journal={arXiv preprint arXiv:2410.11570},
  year={2024}
}

@article{li2025reduce,
  title={Reduce Lap Time for Autonomous Racing with Curvature-Integrated MPCC Local Trajectory Planning Method},
  author={Li, Zhouheng and Xie, Lei and Hu, Cheng and Su, Hongye},
  journal={arXiv preprint arXiv:2502.03695},
  year={2025}
}

@inproceedings{weng2024aggressive,
  title={An aggressive cornering framework for autonomous vehicles combining trajectory planning and drift control},
  author={Weng, Wangjia and Hu, Cheng and Li, Zhouheng and Su, Hongye and Xie, Lei},
  booktitle={2024 IEEE Intelligent Vehicles Symposium (IV)},
  pages={2749--2755},
  year={2024},
  organization={IEEE}
}

@inproceedings{lyons2023curvature,
  title={Curvature-aware model predictive contouring control},
  author={Lyons, Lorenzo and Ferranti, Laura},
  booktitle={2023 IEEE international conference on robotics and automation (ICRA)},
  pages={3204--3210},
  year={2023},
  organization={IEEE}
}

@article{ma2025constraint,
  title={Constraint-Aware Diffusion Guidance for Robotics: Real-Time Obstacle Avoidance for Autonomous Racing},
  author={Ma, Hao and Bodmer, Sabrina and Carron, Andrea and Zeilinger, Melanie and Muehlebach, Michael},
  journal={arXiv preprint arXiv:2505.13131},
  year={2025}
}

@article{song2020denoising,
  title={Denoising diffusion implicit models},
  author={Song, Jiaming and Meng, Chenlin and Ermon, Stefano},
  journal={arXiv preprint arXiv:2010.02502},
  year={2020}
}

@article{chao2022investigating,
  title={On investigating the conservative property of score-based generative models},
  author={Chao, Chen-Hao and Sun, Wei-Fang and Cheng, Bo-Wun and Lee, Chun-Yi},
  journal={arXiv preprint arXiv:2209.12753},
  year={2022}
}

@inproceedings{du2023reduce,
  title={Reduce, reuse, recycle: Compositional generation with energy-based diffusion models and mcmc},
  author={Du, Yilun and Durkan, Conor and Strudel, Robin and Tenenbaum, Joshua B and Dieleman, Sander and Fergus, Rob and Sohl-Dickstein, Jascha and Doucet, Arnaud and Grathwohl, Will Sussman},
  booktitle={International conference on machine learning},
  pages={8489--8510},
  year={2023},
  organization={PMLR}
}

@article{luo2024potential,
  title={Potential based diffusion motion planning},
  author={Luo, Yunhao and Sun, Chen and Tenenbaum, Joshua B and Du, Yilun},
  journal={arXiv preprint arXiv:2407.06169},
  year={2024}
}

@article{chi2023diffusion,
  title={Diffusion policy: Visuomotor policy learning via action diffusion},
  author={Chi, Cheng and Xu, Zhenjia and Feng, Siyuan and Cousineau, Eric and Du, Yilun and Burchfiel, Benjamin and Tedrake, Russ and Song, Shuran},
  journal={The International Journal of Robotics Research},
  pages={02783649241273668},
  year={2023},
  publisher={SAGE Publications Sage UK: London, England}
}

@article{ho2020denoising,
  title={Denoising diffusion probabilistic models},
  author={Ho, Jonathan and Jain, Ajay and Abbeel, Pieter},
  journal={Advances in neural information processing systems},
  volume={33},
  pages={6840--6851},
  year={2020}
}

@article{han2023efficient,
  title={An efficient spatial-temporal trajectory planner for autonomous vehicles in unstructured environments},
  author={Han, Zhichao and Wu, Yuwei and Li, Tong and Zhang, Lu and Pei, Liuao and Xu, Long and Li, Chengyang and Ma, Changjia and Xu, Chao and Shen, Shaojie and others},
  journal={IEEE Transactions on Intelligent Transportation Systems},
  volume={25},
  number={2},
  pages={1797--1814},
  year={2023},
  publisher={IEEE}
}

@article{janner2022planning,
  title={Planning with diffusion for flexible behavior synthesis},
  author={Janner, Michael and Du, Yilun and Tenenbaum, Joshua B and Levine, Sergey},
  journal={arXiv preprint arXiv:2205.09991},
  year={2022}
}

@article{zhang2020optimization,
  title={Optimization-based collision avoidance},
  author={Zhang, Xiaojing and Liniger, Alexander and Borrelli, Francesco},
  journal={IEEE Transactions on Control Systems Technology},
  volume={29},
  number={3},
  pages={972--983},
  year={2020},
  publisher={IEEE}
}

@article{liu2017global,
  title={Global path planning for autonomous vehicles in off-road environment via an A-star algorithm},
  author={Liu, Qinghe and Zhao, Lijun and Tan, Zhibin and Chen, Wen},
  journal={International Journal of Vehicle Autonomous Systems},
  volume={13},
  number={4},
  pages={330--339},
  year={2017},
  publisher={Inderscience Publishers (IEL)}
}

@article{karaman2011sampling,
  title={Sampling-based algorithms for optimal motion planning},
  author={Karaman, Sertac and Frazzoli, Emilio},
  journal={The international journal of robotics research},
  volume={30},
  number={7},
  pages={846--894},
  year={2011},
  publisher={Sage Publications Sage UK: London, England}
}

@article{wu2021st,
  title={ST-FMT*: A fast optimal global motion planning for mobile robot},
  author={Wu, Zheng and Chen, Yanjie and Liang, Jinglin and He, Bingwei and Wang, Yaonan},
  journal={IEEE Transactions on Industrial Electronics},
  volume={69},
  number={4},
  pages={3854--3864},
  year={2021},
  publisher={IEEE}
}

@article{sakcak2019sampling,
  title={Sampling-based optimal kinodynamic planning with motion primitives},
  author={Sakcak, Basak and Bascetta, Luca and Ferretti, Gianni and Prandini, Maria},
  journal={Autonomous Robots},
  volume={43},
  number={7},
  pages={1715--1732},
  year={2019},
  publisher={Springer}
}

@inproceedings{hess2016real,
  title={Real-time loop closure in 2D LIDAR SLAM},
  author={Hess, Wolfgang and Kohler, Damon and Rapp, Holger and Andor, Daniel},
  booktitle={2016 IEEE international conference on robotics and automation (ICRA)},
  pages={1271--1278},
  year={2016},
  organization={IEEE}
}

@article{li6127037evo,
  title={EVO-MPCC: Enhanced Velocity Optimization with Learning-Based Auto-Tuning for Real-Time Vehicle Trajectory Planning},
  author={Li, Zhouheng and Zhou, Bei and Piccinini, Mattia and Hu, Cheng and Zarrouki, Baha and Mangharam, Rahul and Xie, Lei},
  journal={Available at SSRN 6127037},
  year={2026}
}

@article{zhao2026vision,
  title={Vision-Guided MPPI for Agile Drone Racing: Navigating Arbitrary Gate Poses via Neural Signed Distance Fields},
  author={Zhao, Fangguo and Zhang, Hanbing and Li, Zhouheng and Guan, Xin and Li, Shuo},
  journal={arXiv preprint arXiv:2603.07199},
  year={2026}
}

@inproceedings{yamaguchi2021model,
  title={Model predictive path planning for autonomous parking based on projected C-space},
  author={Yamaguchi, Takuma and Ishiguro, Tatsuya and Okuda, Hiroyuki and Suzuki, Tatsuya},
  booktitle={2021 IEEE International Intelligent Transportation Systems Conference (ITSC)},
  pages={929--935},
  year={2021},
  organization={IEEE}
}

@article{wang2024poco,
  title={Poco: Policy composition from and for heterogeneous robot learning},
  author={Wang, Lirui and Zhao, Jialiang and Du, Yilun and Adelson, Edward H and Tedrake, Russ},
  journal={arXiv preprint arXiv:2402.02511},
  year={2024}
}

	\begin{table}[t]
		\centering
		\caption{\revshort{Simulation Evaluation of Models Composition in Out-of-Distribution Scenes.}{Composition Evaluation in Out-of-Distribution Scenes.}}
		\begin{adjustbox}{max width=\columnwidth}
			\footnotesize
			\setlength{\tabcolsep}{2.2pt}
			\renewcommand{\arraystretch}{1.05}
			\begin{tabular}{c|c|c|cc|c|c|cc}
				\toprule
				\multirow{3}[6]{*}{Scenes}   & \multicolumn{1}{c|}{\multirow{3}[6]{*}{Cond}} & \multicolumn{3}{c|}{Uncond-dynamic$^1$}                                              & \multicolumn{1}{c|}{\multirow{3}[6]{*}{Cond}} & \multicolumn{3}{c}{Uncond-static$^1$}                                                                                                                                                                                                                    \\
				\cmidrule{3-5}\cmidrule{7-9} &        & \multirow{2}[4]{*}{F.Rate}          & \multicolumn{2}{c|}{Computation Time}                      &                                            & \multirow{2}[4]{*}{F.Rate} & \multicolumn{2}{c}{Computation Time}                                                                                                                                                        \\
				\cmidrule{4-5}\cmidrule{8-9} &                                               &                                                                                      & Mean                                          & Std                                        &                            &                                                                                      & Mean                                       & Std                                        \\
				\midrule
				\midrule
				\multirow{4}[2]{*}{$CS_1$}   & 0.8                                           & \cellcolor[rgb]{ .949,  .949,  .949} \textcolor[rgb]{ .753,  0,  0}{\textbf{0.00\%}} & \cellcolor[rgb]{ .949,  .949,  .949} 0.643    & \cellcolor[rgb]{ .949,  .949,  .949} 0.100 & 0.3                        & \cellcolor[rgb]{ .949,  .949,  .949} \textcolor[rgb]{ .753,  0,  0}{\textbf{0.00\%}} & \cellcolor[rgb]{ .949,  .949,  .949} 0.618 & \cellcolor[rgb]{ .949,  .949,  .949} 0.002 \\
				& 2.5                                           & 13.92\%                                                                              & 1.039                                         & 1.157                                      & 2.5                        & 53.16\%                                                                              & 1.730                                      & 3.382                                      \\
				& 5.0                                           & 34.18\%                                                                              & 1.800                                         & 3.930                                      & 5.0                        & 43.04\%                                                                              & 2.181                                      & 3.013                                      \\
				& 8.0                                           & 51.90\%                                                                              & 5.630                                         & 30.508                                     & 8.0                        & 32.91\%                                                                              & 5.179                                      & 38.695                                     \\
				\midrule
				\multirow{4}[2]{*}{$CS_2$}   & 0.8                                           & \cellcolor[rgb]{ .949,  .949,  .949} \textcolor[rgb]{ .753,  0,  0}{\textbf{0.00\%}} & \cellcolor[rgb]{ .949,  .949,  .949} 0.653    & \cellcolor[rgb]{ .949,  .949,  .949} 0.011 & 0.2                        & \cellcolor[rgb]{ .949,  .949,  .949} \textcolor[rgb]{ .753,  0,  0}{\textbf{0.00\%}} & \cellcolor[rgb]{ .949,  .949,  .949} 0.622 & \cellcolor[rgb]{ .949,  .949,  .949} 0.019 \\
				& 2.5                                           & 43.04\%                                                                              & 1.446                                         & 2.679                                      & 2.5                        & 68.35\%                                                                              & 2.199                                      & 4.812                                      \\
				& 5.0                                           & 40.51\%                                                                              & 3.554                                         & 11.316                                     & 5.0                        & 64.56\%                                                                              & 4.498                                      & 19.689                                     \\
				& 8.0                                           & 69.62\%                                                                              & 13.163                                        & 91.913                                     & 8.0                        & 53.16\%                                                                              & 6.360                                      & 31.140                                     \\
				\bottomrule
			\end{tabular}%
		\end{adjustbox}
		
		\begin{tablenotes}
			\footnotesize
			\item $^1$ \revshort{The compositional weight of all unconditional diffusion models is 5.}{All unconditional-model weights are 5.}
		\end{tablenotes}
		\label{tab:comp_gen}%
	\end{table}%

\end{document}